\documentclass{article}
\usepackage{spconf,amsmath,graphicx}

\usepackage{booktabs}
\usepackage{mathrsfs}
\usepackage{amssymb}
\usepackage{kotex}
\usepackage{xcolor}
\usepackage{comment}
\usepackage{sansmath}
\usepackage{tikz}

\newcommand{\bg}[1]{\textcolor{black} {#1}} 
\newcommand{\jt}[1]{\textcolor{black}{#1}} 
\newcommand{\bgnew}[1]{\textcolor{black} {#1}} 

\newcommand{\cam}[1]{\textcolor{black} {#1}} 

\newcommand*{\affmark}[1][*]{{\normalfont\textsuperscript{#1}}}

\title{Scalable Weight Reparametrization for Efficient Transfer learning}
\name{Byeonggeun Kim\sthanks{equal contribution}\sthanks{\cam{Work completed during employment at Qualcomm Technologies, Inc.}}, Jun-Tae Lee$^*$\affmark[1], Seunghan Yang\affmark[1], and Simyung Chang\affmark[1]}
\address{\cam{\affmark[1]Qualcomm AI Research\sthanks{Qualcomm AI Research is an initiative of Qualcomm Technologies, \hspace*{2.4em} Inc.}, Qualcomm Korea YH, Seoul, Republic of Korea}
}
\begin{document}
%
\maketitle
\begin{abstract}

\cam{This paper proposes a novel, efficient transfer learning method, called Scalable Weight Reparametrization (SWR) that is efficient and effective for multiple downstream tasks. Efficient transfer learning involves utilizing a pre-trained model trained on a larger dataset and repurposing it for downstream tasks with the aim of maximizing the reuse of the pre-trained model. However, previous works have led to an increase in updated parameters and task-specific modules, resulting in more computations, especially for tiny models. Additionally, there has been no practical consideration for controlling the number of updated parameters. To address these issues, we suggest learning a policy network that can decide where to reparametrize the pre-trained model, while adhering to a given constraint for the number of updated parameters. The policy network is only used during the transfer learning process and not afterward. As a result, our approach attains state-of-the-art performance in a proposed multi-lingual keyword spotting and a standard benchmark, ImageNet-to-Sketch, while requiring \textit{zero} additional computations and significantly fewer additional parameters.}

\end{abstract}
\section{Introduction}
\label{sec:intro}
\vspace{-0.2cm}
Transfer learning has been crucial in various fields, \cam{including} vision, natural language, and audio
\cite{Bert, wav2vec2.0,adapter_icml19}. Conventionally, fine-tuning the entire layers or the last classification layer is widely used to transfer a pre-trained model to a downstream task~\cite{TransferLearning}. The former produces a separate copy of a pre-trained model parameters for each task. Albeit it may achieve higher performance for each task, the parameter efficiency drastically decreases \cam{as} the number of downstream tasks increases, \cam{which can pose a challenge for memory-constrained systems such as mobile phones.}
The latter is more efficient but likely to suffer \cam{from} lower performance.

Hence, efficient transfer learning has been studied to minimize the number of updated parameters and maintain accuracy in downstream tasks~\cite{adapter_icml19, spottune, repnet, input_reprogramming19, cvpr22_taps}.
\cam{However, these approaches~\cite{adapter_icml19, spottune, repnet, input_reprogramming19} usually suggest task-specific extra modules, which increase model complexity. And they commonly suffer from determining the optimal placement and quantity of the extra modules or requiring elaborate hyperparameter tuning to satisfy model constraints regarding the number of updated parameters or computations~\cite{cvpr22_taps}.
Moreover, we have observed that these methods are inefficient, particularly for tiny models targeting edge devices. Thus, it is of practical importance to accurately attain the expected efficiency while ensuring the highest accuracy within a given efficiency cost.}

In this work, we present Scalable Weight Reparametrization (SWR) for achieving efficiency-controllable transfer learning on tiny models. We applies weight reparametrization by \textit{adding} a learnable weight term on each pre-trained weight to obtain \bgnew{a task-specific one, thus needing \textit{zero} extra module and computation}. A policy network determines whether or not to apply the weight reparametrization for \bgnew{each weight} under an efficiency constraint.
\bg{Once the policy is generated, the policy network is no longer used during weight reparametrization.}
Our SWR attains the state-of-the-art (SOTA) performance  \jt{satisfying various expected efficiency levels} for tiny models on a novel multi-lingual keyword spotting benchmark. Moreover, it is effectively extended to a larger backbone on a standard benchmark, ImageNet-to-Sketch~\cite{imagenet-to-sketch_piggyback}. 


\vspace{-0.1cm}
\begin{figure*}[t]
  \centering
  \includegraphics[width=\linewidth]{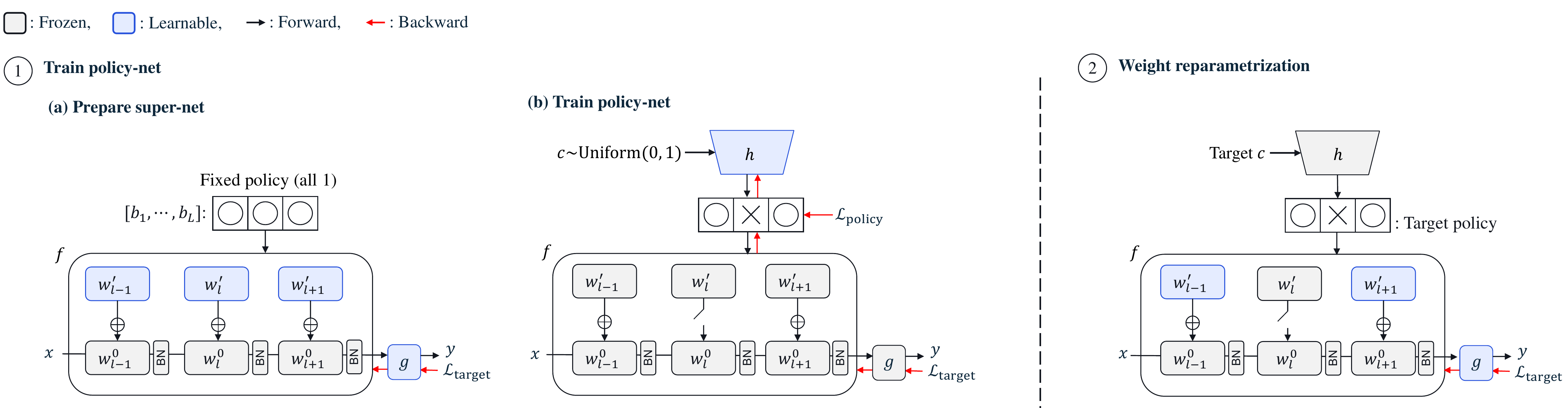}
  \vskip -0.15in
  \caption{\textbf{Two-stage training of SWR.} 
  $f$, $g$, and $h$ are the feature extractor, the classifier, and the policy-net. \jt{Given a downstream task,} $c$ is the target cost for $f$ after transfer learning, and $b$ is the binary policies for $L$ layers. \jt{\raisebox{.5pt}{\textcircled{\raisebox{-.9pt} {1}}}} The policy-net is trained upon the super-net, which fine-tunes all the $L$ layers.
\jt{\raisebox{.7pt}{\textcircled{\raisebox{-.9pt} {2}}}} Layers are updated by weight reparametrization for a target \bgnew{policy}.
  }
  \vskip -0.2in
  \label{fig:overall_system}
\end{figure*}

\section{Related works}
\vspace{-0.2cm}
\noindent\textbf{Efficient transfer learning.} \cam{Several studies, such as \cite{adapter_icml19, spottune, repnet, inter19_google_multi_lingual_ASR}, have suggested task-specific extra modules with pre-trained models.} SpotTune~\cite{spottune} developed instance-specific dynamic gating to decide whether to use a fine-tuned or pre-trained block. Residual Adapter~\cite{adapter_icml19} suggested minimal extra channel-wise operations to adapt to a downstream task. Rep-Net~\cite{repnet} exploited an additional task-specific network parallel to its pre-trained backbone for intermediate feature reprogramming. \cam{In contrast, recently,} TAPS~\cite{cvpr22_taps} suggested a weight reparametrization with learnable scalar thresholding whether to use the reparametrization. While other methods~\cite{spottune, adapter_icml19, repnet} rely on extra task-specific modules, akin to our work, TAPS~\cite{cvpr22_taps} utilized weight reparametrization. However, it is sensitive to hyperparameters and hence hard to optimize for target model efficiency. 
In contrast, we suggest a scalable policy network and an effective training process to achieve both target efficiency and reliably better performances.


\section{Method}
\label{sec:Method}
\vspace{-0.15cm}

\noindent \textbf{\jt{Problem definition.}} 
Given a pre-trained feature extractor, \jt{$f_0$,} there are $T$ downstream tasks with corresponding training datasets, $D_1, D_2, \cdots, D_T$, where $D_t=\{x_i, y_i\}_{i=1}^{N_t}$ with an example $x_i$ and its corresponding label $y_i$. Since the label sets of different tasks do not need to overlap, they usually have different class sets. 
\jt{By transfer learning to a task $t$, $f_0$ is updated to $f_t$, and task-specific classifier $g_t$ is newly learned.
We aim to find $f_t$ and $g_t$ optimized to the corresponding objective and satisfy the constrained amount of updated weights for efficiency at multi-task transferring. Note that, in our work, $f_t$ contains the same number of parameters as $f_0$ (no complexities increase for a single task). We will omit $t$ for brevity.} 

\subsection{Weight reparametrization \& policy network}
\label{subsec:w_reparam}
\vspace{-0.1cm}
\jt{For a learnable layer $l\in \{1, \cdots, L\}$ of $f$,}
freezing the pre-trained weight $w^0_l$, SWR obtains the transferred weight \bg{$w_l$} by a learnable reparametrizing weight $w'_l$ \cam{as~\cite{cvpr22_taps}}:
\begin{equation}
\label{eq:w_reparam}
    w_l = w^0_l + b_l w'_l,
\end{equation}
where $b_l\in\{0,1\}$ is a policy that decides \bgnew{whether} the $l$th layer weight is \bgnew{reparametrized (1) or not (0).} $w'_l$ \bg{is zero-initialized} to let $w_l$ start from $w_l^0$. We reparametrize the learnable layers \bg{of policy 1} in $f$ except for batch normalization layers.

To obtain $b_l$'s, we design a policy network (policy-net) $h$, simply including three linear layers with ReLU \bgnew{in-between}. The policy-net takes the target \bgnew{ratio of additional complexity to the original (refers to target cost, $c\in[0,1]$)}
as an input. \bgnew{The} higher $c$, the more task-specific weights are \bgnew{allowed}. Then, it yields $L$ 2-dimensional outputs, \bg{ $\{h(c)_l\}_{l=1}^{L}$, where each $h(c)_l$'s} first element
\bgnew{, $h(c)_{l,1}$,} is binarized as the policy by a threshold of 0.5 for each $l$. 

\bgnew{During} training, to make the layer-wise binary policy differentiable, we relax the discrete policies to continuous variables with hard Gumbel-Softmax~\cite{gumbel}\bg{, i.e., $b_l = \texttt{Gumbel}(h(c)_l/\tau)_1$, where $\tau$ is softmax temperature and 1 indicates the first element.}

Note that the size of $w_l$ is as same as $w_l^0$, and the policy-net is not required after deployment; hence, there are no increased complexities of the transferred model in terms of the number of parameters and computations in a single downstream task. In multiple $T$ tasks, the increased parameters are only proportional to the amount of task-specific parts, which can be adjusted by $c$.

\noindent\textbf{Transfer of batch normalization (BN).} Previous works~\cite{adapter_icml19, spottune, repnet, cvpr22_taps} commonly update affine transformation parameters as well as running mean \bg{\&} variance in BN layers for each downstream task.
However, their cost is not negligible in tiny networks. Therefore, inspired by \cite{NeurIPS19_TransNorm}, we only update the running mean \bg{\&} variance while 
\jt{freezing} affine parameters to further reduce the number of \jt{updated} parameters.

\subsection{Two-stage training}
\label{subsec:two-stage}
\vspace{-0.1cm}
We observed that co-optimizing $h$ and 
$\{w'_l\}_{l=1}^{L}$ results in sub-optimal convergence. To prevent this, we suggest two-stage training. Fig.~\ref{fig:overall_system} depicts the overall procedure.

\noindent\textbf{(1) Train policy-net with super-net.} 
To get a reliable policy-net, we first learn the target downstream task using a super-net \bgnew{that transfers} the pre-trained model by $w_l = w^0_l + w_l'$, i.e., the policy is 1 for all $l$ in eq.~(\ref{eq:w_reparam}). Hence, the policy-net is not considered here, and the super-net is learned with a target task loss $\mathcal{L}_{\mathrm{target}}$, e.g., conventional cross-entropy loss.

Next, while freezing the super-net, we train a policy-net $h$ varying the target cost input $c\sim \mathrm{Uniform}(0, 1)$.
We devise \bg{an additional} loss term to learn the policy-net as
\begin{equation}
\label{eq:policyloss}
    \mathcal{L}_\mathrm{policy} = | \{\sum_{l=1}^L{r_l\cdot\texttt{Softmax}(h(c)_l/\tau)_1}\} - c|,
\end{equation}
\bg{where \bg{$r_l=\mathcal{N}(w_l)/\sum^L_{i=1}{\mathcal{N}(w_i)}$, a normalized layer-wise weighting according to $\mathcal{N}(w_l)$,} the number of parameters in $w_l$. The $r_l$ makes $\mathcal{L}_\mathrm{policy}$ considering the cost of layer $l$ based on $\mathcal{N}(w_l)$.
Here, we use Softmax instead of Gumbel-softmax in computing $\mathcal{L}_\mathrm{policy}$ \cam{(still use Gumbel-softmax in $\mathcal{L}_\mathrm{target}$)}.}
The total loss is $\mathcal{L} = \mathcal{L}_\mathrm{target} + \lambda\cdot\mathcal{L}_\mathrm{policy}$, \jt{where $\lambda$ is loss weighting hyperparameter.}

\noindent\textbf{(2) Weight reparametrization.} The policy-net is \bgnew{now} frozen and \bg{generates 0 or 1 binarized policies using \bgnew{the} threshold of 0.5 \bgnew{as} in Sec~\ref{subsec:w_reparam}},
given a target cost level $c$. 
Then, we only learn $w'_l$'s whose corresponding policies are 1 with $\mathcal{L}_\mathrm{target}$.


\section{Experiments}
\label{sec:exp}
\vspace{-0.1cm}

\subsection{Experimental settings}
\vspace{-0.1cm}
\textbf{Keyword spotting} (KWS) detects pre-defined keywords in streaming audio~\cite{bcresnet}. We introduce a challenging transfer learning \cam{task that involves} a multi-lingual pre-trained backbone \cam{being adapted} to a monolingual KWS. We assess the averaged accuracy (Acc.$\uparrow$) with a 95\% confidence interval. For efficiency, we report the total number of learnable parameters (\#Param.$\downarrow$) compared to a single backbone and the \bgnew{number of} multiplies (\#Mult.$\downarrow$) for a single downstream task.

\noindent \textbf{Datasets.} 
\cam{We have created a transfer learning benchmark by combining two KWS datasets, Google speech commmands ver. 2 (GSC2) \cite{GSC} for English and Mandarine Chinese Scripted Speech Corpus-Keyword Spotting (MC-KWS)~\cite{MCKWS}. GSC2 comprises 106K utterances from 2.6K speakers with each utterance being 1 sec. long and sampled at 16kHz. We have adapted the standard 12-way classification~\cite{bcresnet} for GSC2. MC-KWS contains 18 Chinese keywords from 37 speakers with each utterance being between [1.3, 6.2] sec. long and sampled at 48kHz. We have excluded the two English keywords, and out of the 18 remaining keywords, 7 are pairs of similar words and 4 are unpaired. We have followed the same preprocessing as in GSC~\cite{GSC, bcresnet} and have also performed 12-way classification. To this end, we have assigned a keyword from each pair and randomly selected one of the unpaired keywords as a class, which we refer to as `unknown words'. The remaining 10 keywords constitute 10 known classes. We have also included the background noises from GSC2 as a separate class `silence', with the length of 2.24 sec. For the train, validation and test sets, we have mapped 6.7k, 1.6k, 1.6k utterances, respectively from 25, 6, 6 speakers.}




\begin{table}[t]
    \caption{\textbf{Top-1 test accuracy and efficiency on} GSC2 and MC-KWS for BCResNet-8 and Res15 backbones (Best: bold-faced, $2^{\textrm{nd}}$: underlined).} 
    \label{table:main_result}
    \centering
    \resizebox{\linewidth}{!}{
    \begin{tabular}{lccccrr}
    \toprule
    Method & Backbone & GSC2 && MC-KWS & \#Param. & \#Mult.\\
    \midrule
    \midrule
    
    Multi-lingual model &  BCResNet-8 & 95.7 (1.6) && 99.3 (0.4) & 1x & 1x \\
    Fine-tuning All & BCResNet-8 & 98.5 (0.1) && 99.8 (0.2) & 3x & 1x\\
    \midrule
    SpotTune~\cite{spottune} & BCResNet-8 & 98.2 (0.3) && 99.7 (0.4) & 4.02x & 1.79x\\
    Adapter~\cite{adapter_icml19} & BCResNet-8 & 98.3 (0.2) && 99.7 (0.2) & 1.43x & 1.53x\\
    Rep-Net~\cite{repnet} & BCResNet-8 & 97.9 (0.5) && 99.7 (0.2) & 2.26x & 2.15x\\
    TAPS~\cite{cvpr22_taps} & BCResNet-8 & 98.2 (0.2) && 99.7 (0.1) & 2.06x & 1x \\
    \midrule
    
    \bg{SWR}, c=0.1 & BCResNet-8 & \underline{98.4 (0.2)} && 99.7 (0.0) & \textbf{1.09x} & 1x\\
    SWR, c=0.3 & BCResNet-8& \textbf{98.5 (0.1)} && \underline{99.8 (0.2)} & 1.33x & 1x\\
    SWR, c=0.5 & BCResNet-8 & \textbf{98.5 (0.1)} &&  \textbf{99.9 (0.1)} & 2.04x & 1x\\
    
    
    \midrule
    \midrule
    
    
    
    
    Multi-lingual model & Res15 & 95.4 (0.8) && 96.2 (2.4) & 1x & 1x \\
    Fine-tuning All & Res15 & 97.7 (0.2) && 99.2 (0.2) & 3x & 1x\\
    \midrule
    SpotTune~\cite{spottune} & Res15 & 96.3 (0.3) && 94.6 (1.3) & 4.08x& 1.54x\\
    Adapter~\cite{adapter_icml19} & Res15 & 97.5 (0.1) && 98.7 (0.5) & 1.22x & 1.03x \\
    Rep-Net~\cite{repnet} & Res15 & 97.5 (0.2) && 98.7 (0.2) & 1.86x & 1.43x \\
    TAPS~\cite{cvpr22_taps} & Res15 & 97.4 (0.2) && 98.6 (0.1) & 2.00x& 1x\\
    \midrule
    SWR, c=0.1 & Res15 & \underline{97.6 (0.1)} && \underline{99.0 (0.5)} & \textbf{1.08x} & 1x\\
    SWR, c=0.3 & Res15 & \underline{97.6 (0.2)} && \textbf{99.3 (0.3)} & 1.54x & 1x\\
    SWR, c=0.5 & Res15 & \textbf{97.7 (0.1)} && \textbf{99.3 (0.4)} & 2.00x & 1x\\
    \bottomrule
    
    \end{tabular}
    }
    \vskip -0.15in
\end{table}

\begin{table}[t]
    \caption{
    \bgnew{\textbf{Ablation studies of SWR on BCResNet-8.}}
    }
    \vskip +0.05in
    \label{table:ablation}
    \centering
    \resizebox{1.\linewidth}{!}{
    \begin{tabular}{lcccccccc}
    \toprule
    &\multicolumn{2}{c}{c = 0.1}&&\multicolumn{2}{c}{c = 0.3}&&\multicolumn{2}{c}{c = 0.5}\\
    \cmidrule{2-3} \cmidrule{5-6} \cmidrule{8-9}
    Method & GSC2 & MC-KWS && GSC2 & MC-KWS && GSC2 & MC-KWS\\
    \midrule
    \midrule
    Not staged (co-optimize) & 97.2 (0.5) & 97.5 (4.7) && 97.9 (0.6) & 99.6 (0.4) && 98.1 (0.3) & 99.6 (0.4)\\
    SWR (Two-stage) & \textbf{98.4 (0.2)} & \textbf{99.7(0.0)} && \textbf{98.5 (0.1)} & \textbf{99.8 (0.2)} && \textbf{98.5 (0.1)} & \textbf{99.9 (0.1)}\\
    \midrule
    ~~+ BN-affine & 98.3 (0.3) & 99.8 (0.1) && 98.3 (0.2) & 99.9 (0.1) && 98.5 (0.1) & 99.9 (0.0)\\
    ~~+ Super-net weight & 98.3 (0.2) & 99.8 (0.1) && 98.4 (0.1) & 99.8 (0.1) && 98.5 (0.1) & 99.8 (0.2) \\
    
    
    \bottomrule
    \end{tabular}
    }
    \vspace{-0.4cm}
\end{table}

\noindent \textbf{Implementation.} 
We employ two tiny backbones: Broadcasting residual networks (BCResNet-8)~\cite{bcresnet} and Res15 \cite{res15}. BCResNet-8 is SOTA on GSC2 through its efficient design with 321k parameters utilizing broadcasted residual learning. Res15 was designed for KWS task based on conventional residual networks~\cite{resnet}. For each, we adopted the data preprocessing and augmentation outlined in the literature and followed the training scheme for a pre-trained backbone, `Multi-lingual model'. \cam{The pre-training involved classification of combined classes in both GSC and MC-KWS.}

We fine-tune 30 epochs for weight reparametrization of our SWR using \bg{$\lambda$ of 1} and Adam optimizer with a learning rate initialized by \{1e-3, 5e-4\} and cosine-annealed without weight decay. 
\bg{Next, we schedule the (Gumbel) Softmax temperature by cosine annealing from 2 to 0.5 \cite{gumbel}.} \bgnew{In the second stage of SWR, we do zero initialization for $w'_l$'s instead of using the super-net for fair comparisons regarding training cost}.
Finally, we select the transferred model at the last epoch for evaluation.


\subsection{Results}
\vspace{-0.2cm}
Table~\ref{table:main_result} demonstrates the KWS results. We compare our SWR with the recent efficient transfer learning methods, Spottune~\cite{spottune}, Residual Adapter~\cite{adapter_icml19}, Rep-Net~\cite{repnet}, and TAPS~\cite{cvpr22_taps}. We follow their official implementations for transfer learning on KWS. The multi-lingual pre-trained backbone shows lower accuracy than a model \cam{entirely} fine-tuned in a single language (`Fine-tuning All') due to the differences in English and Chinese characteristics~\cite{BiEncoder_Eng_Man_interspeech20, weight_factorization_multilingual_interspeech21}.
\cam{Our SWR outperforms the baselines~\cite{spottune,adapter_icml19,repnet,cvpr22_taps}, and even surpasses `Fine-tuning All' which updates the entire layers. Notably, SWR achieves comparable or even better performance than the baselines, merely with the target cost of 0.1, which requires the smallest extra costs, i.e., zero computation and less than 10\% of the number of parameters.}

\begin{figure}[t]
  \centering
  \includegraphics[width=1.\linewidth]{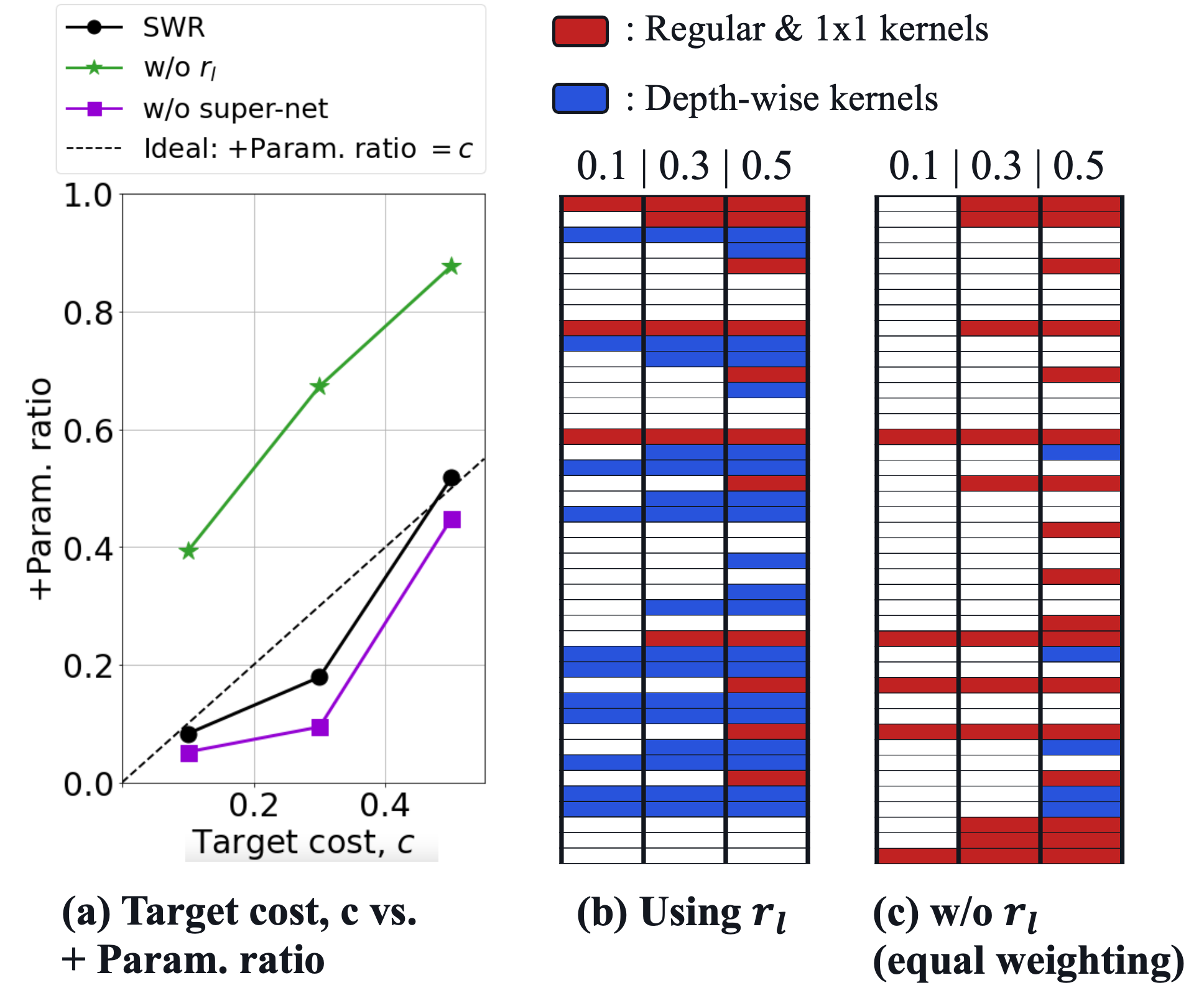}
  \vskip -0.15in
  \caption{\textbf{Controllability of SWR} on GSC2. (a) \jt{Target cost vs. actual increased parameters (+Param. ratio).} (b) \& (c) Resulting policy for 43 convolution layers for different costs \{0.1, 0.3, 0.5\} (layers go deeper from top to bottom).
  }
  \vskip -0.15in
  \label{fig:gating}
\end{figure}

\subsection{Analysis}
\vspace{-0.2cm}
In this section, we analyze SWR on BCResNet-8.

\noindent\textbf{Two-stage training.} \jt{To see the benefit of the \bgnew{two}-stage training, we co-optimize the policy-net $h$ and the reparametrizing weights $w'_l$'s, `Not staged.' The first two rows of Table~\ref{table:ablation} compare it with our complete method, SWR. In all $c$, SWR outperforms `Not staged.' Hence, the \bgnew{two}-stage training is more favorable for optimizing both $h$ and $w'_l$'s.}

\noindent\textbf{Affine paramters in BN \bgnew{\& Super-net weight}.} Comparing the 
\bgnew{second and third}
rows of Table~\ref{table:ablation}, we see that updating affine parameters of BN layers does not guarantee better transfer to downstream tasks but rather inevitably degrades the efficiency in SWR. Contrarily, TAPS~\cite{cvpr22_taps} specifies both layers and affine parameters to each task. Nevertheless, it performs less than the proposed SWR (Table~\ref{table:main_result}). It means that our policy-net finds more informative layers to be updated. \bgnew{Next, the last row in Table~\ref{table:ablation} shows that using super-net, which is fine-tuned with policies of all 1, as initialization of $w_l'$ in the second stage does not guarantee better performance.}

\noindent\textbf{Efficacy of super-net in cost controllablity.}
SWR provides the high-performing transferred model under an efficiency constraint, the target cost $c$.
In Fig.~\ref{fig:gating}(a), we plot the ratio of the increased parameters to the pre-trained model in a transferred model varying $c$, ablating SWR. Hence, the more it matches the ideal (black-dashed), the better controllability. We can see that SWR \bg{(black-circle)} is closer to the ideal than `w/o super-net' \bg{(purple-square)} in all $c$. Compared to SWR, \bg{`w/o super-net' yields undesirable results: less parameters, but degraded performances by 0.1\% for $c$ of 0.1 and 0.3, respectively.} From this, we found that the strategy of super-net-based policy-net learning is essential.

\noindent\textbf{Learned policy.} 
BCResNet-8 includes a single \bg{regular} convolution (conv) and seperated convs where each consists of depth-wise \jt{(temporal and frequency)} and 1x1 \bg{convs}~\cite{bcresnet}. Usually, depth-wise kernels have fewer parameters than the others\bgnew{, hence less weighted by $r_l$ in eq.~(\ref{eq:policyloss})}. Fig.~\ref{fig:gating} shows obtained policies over 43 layers in BCResNet-8 on transfer learning to GSC2 depending on $r_l$. The colored boxes denotes weight-reparametrized layers by policies of 1. Red and blue boxes indicate \{regular conv, 1x1\} and depth-wise kernels, respectively. 
Without $r_l$ \bgnew{(equal weighting)}, the policy-net \jt{is more likely to} select heavy computations (more \bgnew{red boxes} in Fig.~\ref{fig:gating}(c)), resulting in green\bgnew{-star} in Fig.~\ref{fig:gating}(a). On the other hand, with $r_l$, the policy-net prefers the depth-wise kernels to the 1x1 and regular conv kernels as in Fig.~\ref{fig:gating}(b). Then, we can obtain better parameter efficiency (black-circle in Fig.~\ref{fig:gating}(a)).



\begin{table}[t]
    \caption{\textbf{Top-1 test accuracy using ResNet50 on ImageNet-to-Sketch benchmark} (Best: bold-faced, $2^{\textrm{nd}}$: underlined). 
     *: reproduced numbers, ${}^{\dagger}$: results from \cite{cvpr22_taps}}
    \label{table:result_I2S}
    \centering
    \resizebox{\linewidth}{!}{
    \begin{tabular}{lcccccr}
    \toprule
    Method & Flowers & WikiArt & Sketch& Cars & CUB & \#Param. \\
    \midrule
    \midrule
    Fine-tuning All${}^{\dagger}$ & 95.7 & 78.0 & 81.8 & 91.2 & 83.6 & 6x\\
    Fine-tuning Classifier${}^{\dagger}$ & 89.1 & 61.7 & 65.9 & 55.5 & 63.5 & 1x\\
    \midrule
    SpotTune~\cite{spottune} & 96.3 & 75.8 & 80.2 & \textbf{92.4} & 84.0 & 7x\\
    Rep-Net~\cite{repnet} & 96.7 & 76.4* & \underline{80.3}* & 88.7* & 80.3 & 7.85x\\
    TAPS~\cite{cvpr22_taps} & 96.7 & \underline{76.9} & \textbf{80.7} & 89.8 & 82.7 & 4.12x\\
    \midrule
    SWR, c=0.1 & 96.0 & 75.6 & 79.6 & 91.2 & 83.7 & \textbf{1.83x}\\
    SWR, c=0.3 & \underline{96.8} & 76.0 & \underline{80.3} & 91.3 & \underline{84.9} & \underline{2.42x}\\
    SWR, c=0.5 & \textbf{97.0} & \textbf{77.0} & \textbf{80.7} & \underline{92.2} & \textbf{85.2} & 3.40x\\
    \bottomrule
    \end{tabular}
    }
    \vskip -0.2in
\end{table}

\subsection{Extend to larger backbone}
\jt{\noindent\textbf{Benchmark: ImageNet-to-Sketch.} To verify the effectiveness of our SWR in a larger backbone, we apply SWR in a transfer learning benchmark ImageNet-to-Sketch~\cite{imagenet-to-sketch_piggyback},
where} five downstream tasks are Flowers~\cite{Flowers}, CUBS~\cite{CUB}, Cars~\cite{stanfordCars}, Sketch~\cite{sketch}, and WikiArts~\cite{wikiart}, based on a ImageNet~\cite{imagenet} pre-trained model
. 
\jt{In this benchmark, the resnet-50~\cite{resnet} is the standard backbone.} We follow the settings of \jt{the recent related literature and released source code}~\cite{repnet, cvpr22_taps}:
we fine-tune for 50 epochs for SWR using $\lambda=3$, SGD optimizer with a learning rate of \{5e-3, 1e-2, 2e-2\} and cosine annealed with weight decay of \{1e-4, 0\}, \cam{and label smoothing of 0.3}.

\noindent\textbf{Results.} We compare the proposed SWR with the recent methods~\cite{spottune, repnet, cvpr22_taps} in Table~\ref{table:result_I2S}. \bg{Our SWR achieves par or surpasses the SOTA accuracy on four downstream tasks while using fewer additional parameters, 3.40x. Only in the task, Cars, SWR achieves the second best, 92.2, which is lower than 92.4 of SpotTune, \bgnew{which uses} double the additional number of parameters. The results show the feasibility of our SWR for a larger backbone, e.g., resnet-50, at a different domain, Image recognition.} 




\section{Conclusions}
\label{sec:conclusion}
\vspace{-0.1cm}
\cam{In this work, we proposes a novel efficient transfer learning called Scalable Weight Reparametrization (SWR). SWR learns reparametrization of pre-trained weights with the policy network, which determines which layers to reparametrize under an efficiency constraint. We also proposed a two-stage learning process for a reliable training of both reparametrization parameters and the policy network. Regarding accuracy and efficiency, SWR achieved the SOTA in KWS with tiny backbones and successfully extended to a larger backbone in image classification.}

\vfill\pagebreak

\bibliographystyle{IEEEbib}
\bibliography{strings,refs}

\end{document}